\documentclass[a4paper]{article}

\usepackage{INTERSPEECH2021}
 \usepackage{booktabs}
 \usepackage{multirow}
 \usepackage{multicol}
 \usepackage{makecell}
 \usepackage{balance}
 \usepackage{amsthm}
 \usepackage{scalerel}
 \usepackage{cite}
 \usepackage{stackrel}
\title{WavFT: Acoustic model finetuning with labelled and unlabelled data}
\name{}
\name{Utkarsh Chauhan, Vikas Joshi, Rupesh R. Mehta}
\address{Microsoft Corporation}
\email{utkarsh.chauhan, vikas.joshi, rupesh.mehta@microsoft.com}

\begin{document}

\maketitle

\begin{abstract}
Unsupervised and self-supervised learning methods have leveraged unlabelled data to improve the pretrained models. However, these methods need significantly large amount of unlabelled data and the computational cost of training models with such large amount of data can be prohibitively high. We address this issue by using unlabelled data during finetuning, instead of pretraining. We propose acoustic model finetuning (FT) using labelled and unlabelled data. The model is jointly trained to learn representations to classify senones, as well as learn contextual acoustic representations. Our training objective is a combination of cross entropy loss, suitable for classification task, and contrastive loss, suitable to learn acoustic representations. The proposed approach outperforms conventional finetuning with $11.2\%$ and $9.19\%$ word error rate relative (WERR) reduction on Gujarati and Bengali languages respectively. 
\end{abstract}


\section{Introduction}
\label{sec:intro}
Unsupervised learning methods are gaining significant traction in acoustic model training \cite{wav2vec,vqwav2vec,wav2vec2.0,XLSR,WavLM,swietojanski2012unsupervised}. These methods exploit large amount of unlabelled data, which is easily available. For instance, wav2vec\cite{wav2vec} use unlabelled data to pretrain the acoustic models and subsequently finetune with the task specific labelled data. The resultant acoustic models trained by initializing from these pretrained models, show superior performance compared to random initialization. The magnitude of gains is often higher for low resource languages\cite{XLSR, Ai4Bharat}. Therefore an emerging acoustic model training paradigm is to pretrain a seed model with large amount of multilingual unlabelled data and subsequently fine-tune it with language specific labelled data. 


Transfer learning\cite{TL_kunze-etal-2017,TL_2,TL_3,Joshi2020} and multilingual modeling\cite{SHL_1,SHL_2,MLT_1,MLT_2,Arindam_Mrnnt,Multilingual_RNNT,Besacier-ASRUnderResourcedSurvey,joshi2021multiple,LanguageIndependentAM,Lin-MultilingualAM} are also widely used to improve the speech recognition accuracy. It is now a common practice to pretrain a seed model using multilingual labelled data and subsequently finetune with language specific labelled data. While such pretrained models improve the performance, they do not leverage unlabelled data. Alternatively, self-supervised learning\cite{amzn_million,BigSSL} methods use both labelled and unlabelled data for pretraining. These methods use a well trained teacher model to obtain labels or posteriors for the unlabelled data and subsequently use them along with the labelled data to pretrain the acoustic models. Broadly, the  pretrained models can be categorized into following three categories:

\begin{itemize}
    \item Unsupervised seed: Trained with unlabelled data alone using unsupervised learning methods.  
    \item Supervised seed: Trained with labelled data alone using cross entropy loss.
    \item Self-supervised seed: Trained with both labelled and unlabelled data.
\end{itemize}

While unsupervised seed models preform better than random initialization, they are  often found to be inferior compared to supervised seed models, especially in industry scale settings. This is because the labelled data pooled from multiple languages to train the supervised seed model is often large. Self-supervised seed models can perform better than supervised and unsupervised seed models, however, need significantly large amount of unlabelled data compared to labelled data to obtain meaningful gains. \cite{amzn_million} and ~\cite{BigSSL} use close to $1$ million hours of unlabelled data with $7000$ and $34000$ hours of labelled data. Though it is possible to get $1$ million hours of unlabelled data, it is expensive to train models  with such large amount of data. In general, using unlabelled data during pretraining pose the following challenges:

\begin{itemize}
    \item \textbf{Data processing:} Unlabelled data needs processing to extract speech only segments and remove silences, noise, and other sounds. We also need to obtain the labels or posteriors via decoding or forward pass. Hence, processing large amount of unlabelled data can be tedious and computationally expensive.
    \item \textbf{Model training:} Will require large number of compute instances to train models with large amount of unlabelled  data. This cost can be prohibitively high or can severely constrain the experimentation process.
    \item \textbf{Model updates:} Pretrained models are regularly updated with arrival of new data or with new modeling enhancements. Such updates will now be constrained owing to expensive data preparation and model training. 
\end{itemize}

Realizing the above challenges, we propose an acoustic modeling framework which utilizes any available seed model and leverages unlabelled data during finetuning. It needs much smaller amount of unlabelled data, as the amount of labelled data used in finetuning is considerably smaller than used in pretraining. We show improvements with $960$ hours of unlabelled data, when using $500$ hours of  labelled data for finetuning.

 More specifically, we propose to train the model with a joint objective function, consisting of cross entropy loss, suitable for classification task using labelled data, and the contrastive loss, suitable to learn contextual representations using unlabelled data. Therefore, the model is trained to learn representations to classify sound units as well as to learn contextual acoustic representations. The proposed approach is referred to as \textit{WavFT} as it learns contextual acoustic representation during finetuning. We conduct experiments on hybrid automatic speech recognition (ASR) systems and show the efficacy of WavFT on two Indian languages, namely, Gujarati (Gu-IN) and Bengali (Bn-IN). WavFT shows $11.2\%$ and $9.19\%$ WERR reduction over conventional finetuning on Gu-IN and Bn-IN languages, respectively. WavFT is readily applicable to the production scale systems, as the only change in the acoustic model training is during finetuning by using unlabelled data and the accordingly modified objective function. 

The rest of the paper is organized as follows: We discuss prior work in section~\ref{sec:prior_work} and proposed WavFT approach in section~\ref{sec:WavFT}. Experimental set-up is discussed in section~\ref{sec:expt_setup}, followed by results in section~\ref{sec:results}. Finally, conclusions and scope of future work is discussed in section~\ref{sec:conclusion}.

\section{Relation to prior work}
\label{sec:prior_work}
Transfer learning \cite{TL_kunze-etal-2017,TL_2,TL_3,Joshi2020} leverage a well trained seed model from high resource language to improve the accuracy of the low resource language. A natural extension to transfer learning is to train multilingual seed models\cite{SHL_1,SHL_2,MLT_1,MLT_2,Arindam_Mrnnt,Multilingual_RNNT,Besacier-ASRUnderResourcedSurvey,joshi2021multiple,LanguageIndependentAM,Lin-MultilingualAM} with data pooled from multiple languages. While both these methods are effective, they do not leverage easily available unlabelled data.

Recently, numerous studies have used unsupervised and self-supervised learning  to improve the speech recognition performance, especially on low resource languages. Wav2vec\cite{wav2vec} show the efficacy of pretraining the model to learn contextual acoustic representations using unlabelled data. Vqwav2vec\cite{vqwav2vec} introduce learning discrete vector representations. Wav2vec2.0\cite{wav2vec2.0} further improve by masking the speech input in the latent space and solving a contrastive task defined over quantized latent representations. XLSR\cite{XLSR} model showed significant improvements on low resource languages by pretraining wav2vec2.0 with multilingual data. The effectiveness of such pretrained multilingual models on Indian languages is shown in \cite{Ai4Bharat}. While the pretrained models trained with above discussed unsupervised learning methods significantly improved over random initialization, their performance is often inferior to supervised seed initialization, as they do not use labelled data during pretraining. Authors in \cite{amzn_million} show improvements with semi-supervised learning using $1$ million hours of unlabelled data. A detailed study of unsupervised and self-supervised methods for pretraining is done in \cite{BigSSL} and authors show benefits of such pretrained models on numerous downstream tasks. UniSpeech \cite{UniSpeech} proposed learning unified speech representations using labelled and unlabelled data during pretraining, and showed improvements with such pretrained models on down-stream tasks. All the above discussed methods use unlabelled data during pretraining stage, unlike our proposition to using it in finetuning for said advantages. Also, our method can improve on top of these methods by using the corresponding pretrained models. 


\section{WavFT}
\label{sec:WavFT}
During finetuning, the acoustic model learns representations to classify senones using labelled data. Often such labelled data is small and can lead to over-fitting and poor generalization. On the other hand,  model can learn robust contextual acoustic representations with unlabelled data using unsupervised learning\cite{wav2vec,wav2vec2.0} methods. We hypothesize that learning contextual acoustic representations along with representations to classify senones, can help model generalize better.  We achieve this by training the model with a joint objective function consisting of: a) The cross entropy loss to learn  representations for classification task, computed on labelled data only b) The contrastive loss to learn acoustic representations for better generalization, computed on both labelled and unlabelled data. We next discuss the model architecture and training objective in detail.

\begin{figure}[th]
\includegraphics[scale=0.55]{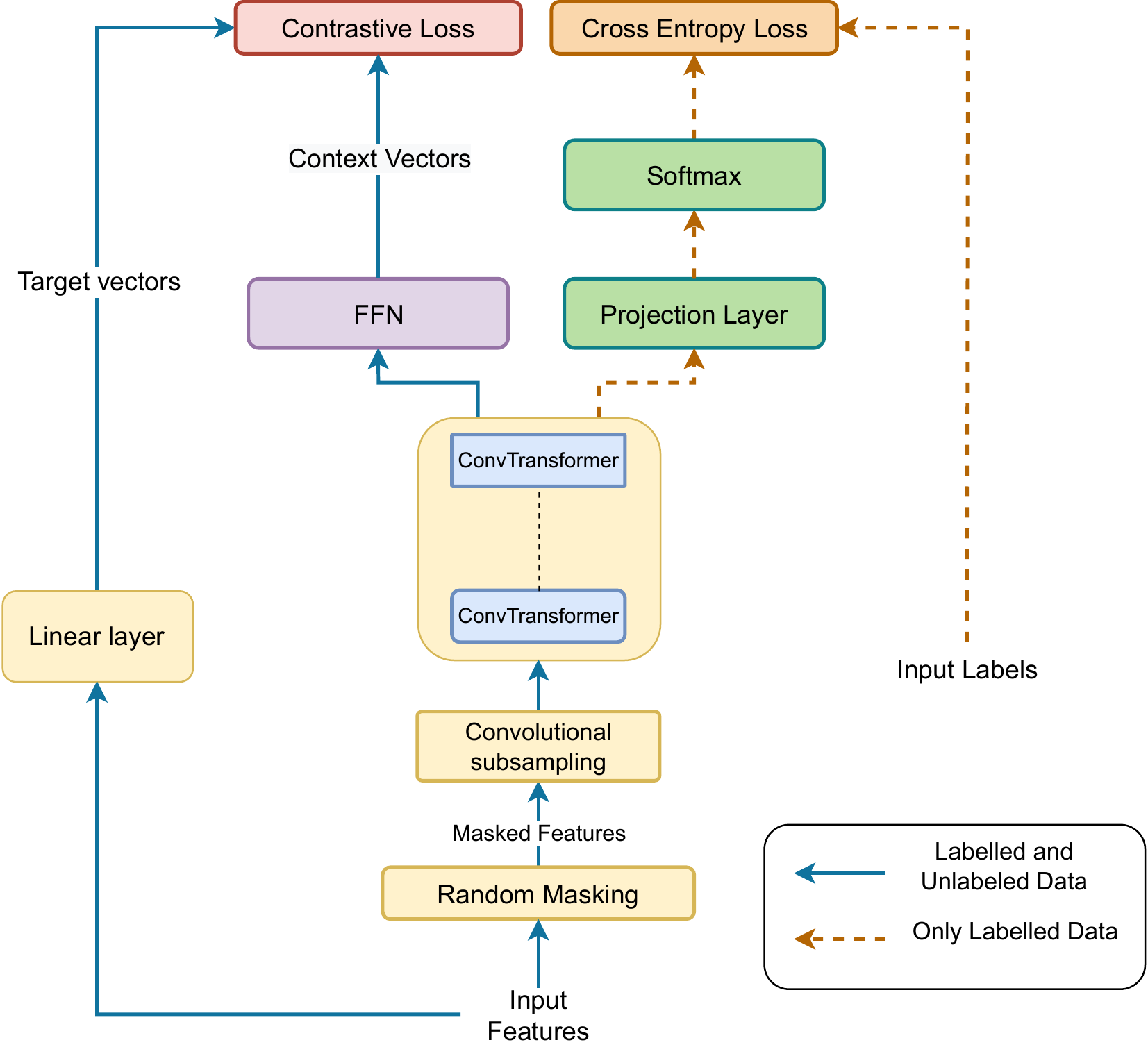}
\caption{The wavFT finetuning.}
\label{fig:WavFT}
\end{figure}

\textbf{Model architecture: }Fig.~\ref{fig:WavFT} depicts the model architecture and the  proposed approach to finetune the acoustic model with  labelled and unlabelled data. The acoustic model consists of $18$ convolutional transformer (convTransformer)  \cite{convTransformer_speech,convTransformer_vision} blocks. Each block consists of multi-head attention\cite{Transformer_Attn} with relative positional embedding,  depth-wise convolution  and feed forward neural (FFN) network modules. The input  log-mel filter bank (LFB) features are randomly masked and passed through the convolutional sub-sampling block. The corresponding features are fed to the convTransformer model to produce latent representations. These representations are further projected to the label dimension using a linear projection layer and subjected to softmax to obtain the posterior probabilities over the labels, senones in our case. The cross entropy loss is computed between the input labels and posterior probabilities only for the labelled data as shown in Fig.~\ref{fig:WavFT}. These latent representations from convTransformer model are also passed through a feed forward neural (FFN) network  to produce context vectors, $c_t$. The LFB features are transformed with a linear layer, instead of quantization, to produce target vectors, $q_t$. The contrastive loss is computed between the context and target vectors for both labelled and unlabelled data. 

\textbf{Training objective:} Let $\mathbb{L}=\{B_{1}^{L},B_{2}^{L},\ldots B_{M}^{L}\}$ represent $M$ batches of labelled data. Let $(\boldsymbol{x},\boldsymbol{y})$ represent an utterance with audio-label pairs in batch $B^L_i$. The cross entropy loss, $\mathcal{L}_{CE}$, computed on utterance $(\boldsymbol{x},\boldsymbol{y})$ is as shown below.

\begin{equation}
\mathcal{L}_{CE}=\frac{-1}{T}\stackrel[t=1]{T}{\sum}\stackrel[c=1]{C}{\sum}y_{t,c}\log \hat{y}_{t,c}
\end{equation}

where $T$ represents the total number of frames in utterance $(\boldsymbol{x},\boldsymbol{y})$. $C$ represents the label dimension. $\boldsymbol{y}_t$ is the one-hot vector representing the output label at time, $t$. $y_{t,c}$ represents the actual value of  $c^{th}$ label at time $t$. $\hat{y}_{t,c}$ is the posterior probability of $c^{th}$ label obtained at time $t$ by processing the input frame $\boldsymbol{x}_{t}$ with the acoustic model.

Let $\mathbb{U}=\{B_{1}^{U},B_{2}^{U},\ldots B_{N}^{U}\}$ represent $N$ batches of unlabelled data. Each batch consists of only audios. Each audio utterance is processed through the acoustic model to produce context and target vectors for every time instant. Given the context vector, $c_t$, the model is trained to identify the right target vector, $q_t$, from a set of candidate representations $\tilde{q}\sim Q_{t}$ using the contrastive loss, $\mathcal{L}_{C}$, defined below:

\begin{equation}
\mathcal{L}_{C}=-\log\frac{\exp(sim(c_{t},q_{t})/k)}{\sum_{\tilde{q}\sim Q_{t}}\exp(sim(c_{t},\tilde{q})/k)}
\end{equation}

where $sim(a,b)$ is the cosine similarity between the two vectors. The computation of contrastive loss is same as done in wav2vec2.0\cite{wav2vec2.0}, except that we use linear layer instead of quantization as done in \cite{BigSSL}.  During training, a labelled or unlabelled batch is selected with probability $p$ and $1-p$, respectively. The final loss depends on the type of the selected batch as defined below:


\begin{equation}
\mathcal{L=}\begin{cases}
\alpha \mathcal{L}_{CE}(B_{i})+(1-\alpha)\mathcal{L}_{C}(B_{i}) & B_{i}\in B^{L}
\mathcal{L}_{C}(B_{i}) B_{i}\in B^{U}
\end{cases}
\label{eqn:WavFT}
\end{equation}


where $\mathcal{L}_{CE}(B_{i})$ and $\mathcal{L}_{C}(B_{i})$ represent the cross entropy and contrastive loss computed on the entire batch of data. $B^L$ and $B^U$ represent set of labelled and unlabelled batches, respectively. Therefore, if the selected batch is labelled, then the final loss is a weighted combination of cross entropy and contrastive loss. If the selected batch is unlabelled, then the final loss is only the contrastive loss. The hyper-parameter $\alpha$ determines the weight between the cross entropy and contrastive loss. During inference, we do not mask the feature and only use convTransformer model along with the projection layer, ignoring the rest of the model blocks.

\section{Experimental details}
\label{sec:expt_setup}
\textbf{Data:} We conduct experiments on $2$ Indian languages, namely Bengali (Bn-IN) and Gujarati (Gu-IN), prominent languages spoken in east and west part of India, respectively. We use $500$ hours of Gujarati and $465$ hours of Bengali labelled data for finetuning. Approximately $20000$ hours of labelled data from seven Indian languages namely, Indian English, Hindi, Gujarati, Tamil, Telugu, Bengali and Marathi is used for supervised seed model training. We use $960$ and $2300$ hours of Gujarati and Bengali unlabelled data, respectively. Unlabelled data is extracted from videos containing tutorials, news broadcasts and continuous conversations such that they largely contain speech segments. Evaluation set consists of $6164$ Gujarati and $9475$ Bengali utterances from different scenarios such as dictation, call center, conversational speech and voice commands.    

\textbf{Experimental set-up:} We conduct experiments on conventional hybrid ASR system consisting of separate acoustic model, language model and lexicon. $80$-dimensional log mel filter bank (LFB) features are computed every $10$ milliseconds (ms). The adjacent LFB features are concatenated to obtain $160$-dimensional features. They are further sampled with a sampling factor of $2$ to obtain $160$-dimensional feature vector for every $20ms$. The acoustic model consists of $18$ convolutional transformer blocks. Each block consists of multi-head attention with relative positional embedding, convolutional block and FFN network. Multi-head attention uses $8$ attention heads with inner dimension of $624$. The  convolutional module use kernel size of $3$ and the FFN network has dimension of $2048$. We use Adam optimizer with linear warm-up of learning rate for $10\%$ of data followed by linear decay for the rest of the data. Unlabelled data is processed through voice activity detection to obtain speech only segments, which are further processed to obtain $160$ dimensional LFB features for every $20ms$. The 5-gram Gujarati and Bengali language model is trained on Gujarati and Bengali text corpus respectively. The labelled batch sampling probability $p$ is set to $0.5$ in all the experiments.  

\section{Discussion of results}
\label{sec:results}
The discussion of results is organized as follows: We first compare different initialization methods and select the best initialization method for further experiments.  Next, we discuss results for WavFT and compare it with the corresponding baseline. We then discuss the results of tuning the hyperparameter, $\alpha$, that decides the importance between cross entropy and contrastive loss. We then discuss our experiments on varying the  amount of unlabelled data and discuss the observations. We finally discuss the impact of using language specific versus multilingual unlabelled data during finetuning. 

\subsection{Comparison of initialization methods}
\label{sec:comp_init}



Table~\ref{tab:seed_init} depicts the WER for acoustic models finetuned with the following three initialization methods namely: a) Random initialization b) Unsupervised seed initialization where the seed model is trained with unlabelled data only using contrastive loss in wav2vec2.0 fashion\cite{wav2vec2.0} c) Supervised seed initialization where seed model is trained with approximately $20000$ hours of multilingual labelled data using cross entropy loss. The  acoustic model for respective initialization method is trained by finetuning the initialized model with corresponding labelled data. The Gu-IN and Bn-IN models are finetuned with $500$ and $465$ hours of corresponding labelled data. As seen from Table~\ref{tab:seed_init}, unsupervised seed initialization shows $10.9\%$ and $18.5\%$ word error rate relative (WERR) reduction compared to random initialization on Gu-IN and Bn-IN locales, respectively. The supervised seed initialization has lower WER compared to other two initialization methods with $18.23\%$ WERR reduction over random initialization on Gu-IN. It also shows $13.9\%$ WERR reduction on Bn-IN locale compared to random initialization. Though unsupervised seed model performs better than supervised seed model initialization in Bn-IN, we still use supervised seed model initialization for both Bn-IN and Gu-IN for sake of consistency. Note that WavFT is agnostic to initial seed model initialization and any seed model can be used. Other methods like self-supervised seed initialization can also be used for WavFT finetuning, however, we do not experiment with them as the main focus of this work is to use unlabelled during finetuning. 

\subsection{WavFT results}
\label{sec:res_WavFT}



\begin{table}
\begin{tabular}{|c|c|c|c|}
\hline 
 & Random init. & Unsupervised seed & Supervised seed\tabularnewline
\hline 
\hline 
Gu-IN & 17.33 & 15.44 & 14.17\tabularnewline
\hline 
Bn-IN & 26.42 & 21.90  & 22.73 \tabularnewline
\hline 
\end{tabular}\caption{WER$[\%]$ comparison for random, unsupervised and supervised seed initialization. The Gu-IN and Bn-IN AMs are  finetuned with the corresponding labelled data after initializing with respective seed models.}
\label{tab:seed_init}
\end{table}

\begin{table}
\begin{tabular}{|c|c|c|}
\hline 
 & \thead{Baseline \\(FT with labelled data)} & \thead{WavFT\\(FT with labelled and \\ unlabelled data)}\tabularnewline
\hline 
\hline 
Gu-IN & 14.17 & 12.59\tabularnewline
\hline 
Bn-IN & 22.73 & 20.65\tabularnewline
\hline 
\end{tabular}\caption{WER$[\%]$ comparison between conventional finetuning (Baseline) and the proposed WavFT approach.}
\label{WavFT}
\end{table}


Table~\ref{WavFT} depicts WER for conventional finetuning (baseline) and the proposed WavFT  for Gu-IN and Bn-IN locales. The baseline Gu-IN and Bn-IN AMs are trained by finetuning the supervised seed model with Gu-IN and Bn-IN labelled data, respectively. The corresponding WavFT AMs are trained by finetuning the same supervised seed model with both labelled and unlabelled data using WavFT approach discussed in section~\ref{sec:WavFT}. WavFT shows $11.2\%$ and $9.19\%$ WERR reduction over conventional finetuning on Gu-IN and Bn-IN locales respectively. The hyperparameter $\alpha$ is set to $0.75$ for Gu-IN and $0.5$ for Bn-IN as they showed the best results. It is not clear why the magnitude of gain is higher for Gu-IN, even though the amount of unsupervised data is more for Bn-IN. We will investigate this further in future work. We next discuss our experiments on hyperparameter tuning, using different amount of unlabelled data and using multilingual unlabelled data. We conduct these experiments on Bn-IN locale as that has more unlabelled data.  


\subsection{Results with varying $\alpha$}
~\label{sec:alpha_0}

\begin{table}
\begin{tabular}{|c|c|c|c|c|c|}
\hline & $\alpha$=0.05 & $\alpha$=0.25 & $\alpha$=0.5 & $\alpha$=0.75 & $\alpha$=1\tabularnewline
\hline 
\hline Bn-IN & 22 & 20.82 & 20.65 & 21.12 & 21.03\tabularnewline
\hline 
\end{tabular}\caption{WER$[\%]$ comparison for different values of $\alpha$. }
\label{tab:vary_alpha}
\vspace{-0.5cm}
\end{table}

The hyper-parameter $\alpha$ determines the weight between cross entropy and contrastive loss in the training objective of WavFT approach, as defined in Eqn.~\ref{eqn:WavFT}. 
$\alpha=0$ implies using only contrastive loss for both labelled and unlabelled data, equivalent to finetuning with unlabelled data alone. As expected, the WER  with $\alpha=0$ is very high ($>95\%$) as the model forgets to classify senones and only learns acoustic representations. $\alpha>0.5$ implies giving more importance to classification task rather than to learn contextual acoustic representations and vice versa.  Table~\ref{tab:vary_alpha} shows WER with varying values of $\alpha$ on Bn-IN locale. The lowest WER is seen for $\alpha=0.5$ indicating that a combination of cross entropy and contrastive loss is better for labelled data. $\alpha=1.0$ implies using only cross entropy loss for labelled data and contrastive loss only for unlabelled data. This is identical to conventional finetuning, except for using unlabelled data and the corresponding loss applied on the unlabelled data. With $\alpha=1.0$, we see $7.4\%$ WERR reduction compared to conventional finetuning, thereby showing the importance of using unlabelled data. 



\subsection{Results with varying amount of unlabelled data}

\begin{table}
\begin{tabular}{|c|c|c|c|c|}
\hline 
 & $\beta=0.1$ & $\beta=0.5$ & $\beta=1$ & $\beta=5$\tabularnewline
\hline 
\hline 
Bn-IN & 21.71 & 21.41 & 21.36 & 21.03\tabularnewline
\hline 
\end{tabular}\caption{WER$[\%]$ comparison for Gu-IN acoustic models trained with different amount of unlabelled data. $\beta$ represents the ratio of number of hours of unlabelled data to number of hours of labelled data.}
\label{tab:vary_data}
\vspace{-0.5cm}
\end{table}

We conduct experiments on Bn-IN language with varying amount of unlabelled data while keeping the labelled data fixed to $465$ hours. Let $\beta$ represent the ratio of number of hours of unlabelled data to number of hours of labelled data. Table~\ref{tab:vary_data} shows WER with different values of $\beta$. $\beta=1$ implies using $465$ hours of unlabelled data and $\beta=5$ corresponds to using entire unlabelled data, as the unlabelled data is roughly $5$ times our labelled data. For every value of $\beta$, the necessary number of hours of unlabelled data is randomly sampled from the entire set. We use $\alpha=1.0$ for these experiments, as any smaller value of $\alpha$ would imply using all the audios from labelled data to learn acoustic representations and may not truly reflect the impact of different amounts of unlabelled data. Hence, the WER for $\beta=5$ match the WER with $\alpha=1$ in Table~\ref{tab:vary_alpha}. WER decreases with increase in the unlabelled data, however, the amount of gains are not significant after $\beta=0.1$. We did not expect improvements with $47$ hours of unlabelled data ($\beta=0.1$) and also we expected more improvements with increase in the unlabelled data. A possible reason for observing good gains even with smaller unlabelled data is because we kept the labelled batch sampling probability, $p=0.5$, for all values of $\beta$. Therefore, the model is updated with similar number of labelled and unlabelled batches at any time, and even $47$ hours of unlabelled data ($\beta=0.1$) could be diverse enough to learn the acoustic representations.   

\subsection{WavFT results with locale specific and multilingual unlabelled data}

Table~\ref{tab:locale_multi} shows results for Bn-IN AMs finetuned in WavFT fashion with locale specific and multilingual unlabelled data. The locale specific WavFT model is trained with $2300$ hours of Bn-IN unlabelled data. The multilingual WavFT model is trained $9800$ hours of unlabelled data collected from seven Indian languages. $2300$ hours of Bengali data is part of the $9800$ hour of multilingual unlabelled data. In both the models, same $465$ hours of labelled data is used. The batch sampling probability, $p$ and hyperparameter $\alpha$ are set to $0.5$ for both models. WER results in table~\ref{tab:locale_multi} suggest that it is better to use locale unlabelled data instead of multilingual unlabelled data.


\begin{table}
\begin{tabular}{|c|c|c|}
\hline 
 & Locale specific & Multilingual\tabularnewline
\hline 
\hline 
Bn-IN & 20.65 & 21.08\tabularnewline
\hline 
\end{tabular}

\caption{WER$[\%]$ comparison of Gu-IN AMs trained with locale specific and multilingual unlabelled data.}
\label{tab:locale_multi}
\end{table}

\section{Conclusions}
\label{sec:conclusion}

In this work, we propose WavFT, acoustic model finetuning approach with labelled and unlabelled data. The method encompasses selecting a labelled or unlabelled  batch with probability $p$ and updating the model with an objective function that is a weight combination of cross entropy and contrastive loss for labelled batch, and only contrastive loss for unlabelled batch. Our approach shows $11.2\%$ and $9.19\%$ WERR reduction over conventional finetuning on Gu-IN and Bn-IN locales respectively. WavFT needs smaller amount of unlabelled data unlike most unsupervised and self-supervised learning methods which use them in pretraining and hence is computationally less expensive. Our approach can leverage any existing seed model. It is readily usable in the production scale acoustic models as the proposed changes differ from conventional finetuning only in terms of using unlabelled data and the accordingly modified objective function. In future, we will conduct experiments with sampling probability $p$, will explore newer unsupervised objective functions, newer model architectures and test the efficacy on end-to-end ASR systems. 

\clearpage
\newpage
\bibliographystyle{IEEEtran}
\balance
\bibliography{main}


\end{document}